\documentclass[conference]{IEEEtran}
\IEEEoverridecommandlockouts
\usepackage{cite}
\usepackage{amsmath,amssymb,amsfonts}
\usepackage{algorithmic}
\usepackage{graphicx}
\usepackage{textcomp}
\usepackage{xcolor}
\usepackage[colorinlistoftodos]{todonotes}
\usepackage{graphicx}
\usepackage{subcaption}
\usepackage{booktabs}
\usepackage{comment}
\usepackage{lineno}
\usepackage{colortbl}
\usepackage[colorlinks=true, linkcolor=black, urlcolor=blue, citecolor=black]{hyperref}
\def\BibTeX{{\rm B\kern-.05em{\sc i\kern-.025em b}\kern-.08em
    T\kern-.1667em\lower.7ex\hbox{E}\kern-.125emX}}
\begin{document}
\title{HOIverse: A Synthetic Scene Graph Dataset With Human Object Interactions
}

\author{
\IEEEauthorblockN{Mrunmai Vivek Phatak}
\IEEEauthorblockA{\textit{Machine Learning and Computer Vision Lab} \\
\textit{Universität Augsburg}\\
Augsburg, Germany \\
mrunmai.phatak@uni-a.de}
\and
\IEEEauthorblockN{Julian Lorenz}
\IEEEauthorblockA{\textit{Machine Learning and Computer Vision Lab} \\
\textit{Universität Augsburg}\\
Augsburg, Germany \\
julian.lorenz@uni-a.de}
\and
\IEEEauthorblockN{Nico Hörmann}
\IEEEauthorblockA{\textit{Machine Learning and Computer Vision Lab} \\
\textit{Universität Augsburg}\\
Augsburg, Germany \\
nico.hoermann@uni-a.de}
\and
\IEEEauthorblockN{Jörg Hähner}
\IEEEauthorblockA{\textit{Organic Computing Lab} \\
\textit{Universität Augsburg}\\
Augsburg, Germany \\
joerg.haehner@uni-a.de}
\and
\IEEEauthorblockN{Rainer Lienhart}
\IEEEauthorblockA{\textit{Machine Learning and Computer Vision Lab} \\
\textit{Universität Augsburg}\\
Augsburg, Germany \\
rainer.lienhart@uni-a.de}
}

\maketitle

\begin{abstract}
When humans and robotic agents coexist in an environment, scene understanding becomes crucial for the agents to carry out various downstream tasks like navigation and planning. Hence, an agent must be capable of localizing and identifying actions performed by the human. Current research lacks reliable datasets for performing scene understanding within indoor environments where humans are also a part of the scene.
Scene Graphs enable us to generate a structured representation of a scene or an image to perform visual scene understanding. To tackle this, we present HOIverse a synthetic dataset at the intersection of scene graph and human-object interaction, consisting of accurate and dense relationship ground truths between humans and surrounding objects along with corresponding RGB images, segmentation masks, depth images and human keypoints. We compute parametric relations between various pairs of objects and human-object pairs, resulting in an accurate and unambiguous relation definitions. In addition, we benchmark our dataset on state-of-the-art scene graph generation models to predict parametric relations and human-object interactions. Through this dataset, we aim to accelerate research in the field of scene understanding involving people.
\end{abstract}


\section{Introduction}
\begin{figure}
    \centering
    \includegraphics[width=9cm]{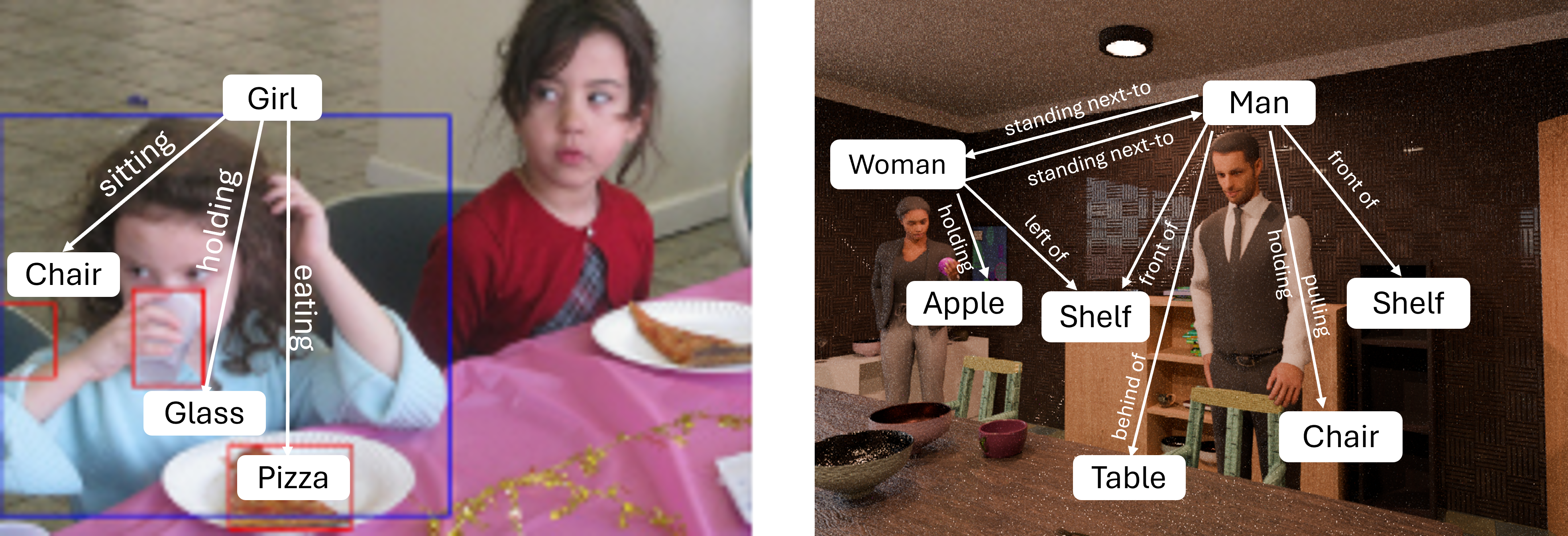}
    \caption{A comparison between annotations provided by VCOCO\cite{gupta2015visual} and HOIverse dataset: VCOCO dataset (left image) annotates only salient interactions, whereas in HOIverse dataset (right image) not only human object interactions are provided but also spatial relations between human and surrounding objects are annotated.}
    \label{fig:vcoco_comparison}
\end{figure}

When humans and robots co-exist in an environment, scene understanding becomes crucial to perform downstream tasks like robot navigation\cite{singh2023scene}, reasoning\cite{hildebrandt2020scene} and planning\cite{gu2024conceptgraphs}. In such scenarios, recognizing not only humans but also their interactions with surrounding objects is important. Scene graphs \cite{Rk_definition} provide a powerful way of representing this information in a structured manner by defining interactions or relations in the form of \textit{(subject, relation, object)} triplets. A subject is either a human or any scene object. With such a triplet, we can describe different relations in the scene. For example, in Fig.\ref{fig:vcoco_comparison} (right image), we can describe human interactions with triplets \textit{(human, holding, apple)}, \textit{(human, behind of, table)}, thus allowing us to store the scene description in a compact form.

Existing scene graph datasets do not explicitly focus on human-object interactions which, however, frequently occur in household settings. Furthermore, traditional scene graph datasets \cite{visual_genome},\cite{psg} are manually annotated, primarily focusing on the visually most salient relations in an image, resulting in sparse scene graph annotations. In case of spatial relations like \textit{front} or \textit{behind} or \textit{next to}, interpretation may vary from person to person, hence making them inconsistent. Therefore, relying on such scene graph datasets would lead to incomplete and inaccurate scene understanding.

Similarly, conventional human-object interaction datasets\cite{gupta2015visual} are manually labeled, extracting the most prominent interactions from available textual descriptions. Therefore, they mainly consist of relations where humans are in direct contact with the surrounding object instances. This results in incomplete annotations that miss many less obvious relations between a human and other objects in a scene. For understanding a scene completely, a broader perspective is needed beyond salient information. For instance, consider the left image in Fig.\ref{fig:vcoco_comparison} taken from VCOCO \cite{gupta2015visual} dataset. While it captures the prominent human-object interactions, understanding the broader scene – whether the girl is sitting \textit{next-to} anybody or where she is sitting – requires considering information not typically annotated, yet crucial for a complete scene interpretation. 

 
Human-object interaction (HOI) closely relates to scene graph generation as both focus on extracting relations. Therefore, in this paper, we present a synthetic scene graph dataset at the intersection of scene graph generation and human-object interaction to tackle the problem of scene understanding by providing complete and consistent scene graph annotations. Our HOIverse dataset primarily focuses on indoor scenes including humans interacting with all kinds of objects. These interactions are modeled as relations. We compute the set of relations between all subject-object instances (including humans) in a procedural way (see Fig.\ref{fig:vcoco_comparison}   right image), without requiring any manual annotations. This results in complete and consistent scene graph annotations.

We develop here a pipeline for adding humans at different locations and in diverse poses in 3D indoor scenes to simulate human-object interactions. This automatic pipeline populates 3D scenes with humans and generates interaction annotations in a procedural way. For every underlying 3D scene, we generate scene graph annotations for all human-object pair and all subject-object pair for multiple camera locations. 
We also include a camera view for each person in the scene, depicting first-person perspective. Often ambiguity is associated while defining spatial relationships like \textit{next to, front, behind}, etc., therefore, parametric relations with additional parameters are computed for the exact interpretation of these relations. Parametric relations are first introduced in \cite{infinisg}, which provide a novel way of extending predicate class definition with parameters such as distance and angles. Based on this idea, we compute parametric relations between all instances of the objects and subjects (including humans) within the scene.
Additionally, for each human instance, we introduce a novel set of parametric relations including \textit{looking at, pointing at} and \textit{body facing} to support underlying human-robot collaboration application.

HOIverse is the first synthetic scene graph dataset that describes human object interactions and presents accurate ground truths specifically for indoor 3D scenes. We provide extensive and accurate relations along with accurate ground-truth RGB images, segmentation masks, depth images and human keypoints. In addition to this, we also provide a comparison between established scene graph generation models (including MotifNet\cite{motifs}, VCTree\cite{VCTree} and DSFormer\cite{dsformer}) on HOIverse dataset on learning to predict human-object interactions and parametric relations. The contributions of this paper can be summarized as follows:
\begin{enumerate}
    \item We present a dataset with exhaustive and fine grained annotations, encoding relationships between humans and surrounding objects within household settings.
    \item We provide a pipeline for adding people with various predefined poses to 3D scenes and generating accurate scene graph annotations, RGB images, segmentation masks, depth images and human keypoints. 
    \item We integrate multiple camera views including a separate camera view per human within each scene to provide first-person perspective additionally, resulting on average in 1.35k relations per camera-view and 22M relations for the entire dataset.
    \item We compare the performances of existing scene graph models on our dataset to predict human-object interactions and parametric relations.
\end{enumerate}

The dataset is publicly available and the latest updates can be found at website: \url{https://mrunmaivp.github.io/hoiverse/}.




\section{Related Work}
Our HOIverse dataset is targeted to provide a reliable scene graph dataset for performing scene understanding within indoor environments with humans. 
This section provides an overview about most widely used scene graph and HOI datasets and how our dataset differentiates from them. 


\subsection{HOI Datasets}

Traditional human-object interaction datasets focus on the task of action recognition. These datasets are manually annotated with bounding boxes around the person performing the action \cite{everingham2010pascal, gkioxari5212r, andriluka20142d}. However, such datasets perform action recognition by only annotating a person and do not provide any annotation for the object of interactions. This results in incomplete visual understanding.


VCOCO \cite{gupta2015visual} dataset introduced Visual Semantic Role Labeling by annotating people instances and their object interactions with corresponding semantic roles. This dataset is derived from the MS-COCO \cite{coco} dataset, by focusing on frequently occurring actions. However, annotations, performed by means of Amazon Mechanical Turk (AMT), consider only salient people (occupying most of the image pixel area) and a binary label per action category. Our work, in contrast not only focus on salient people, but also captures all human instances in the background along with their interactions.


The Trento Universal Human Object Interaction (Tuhoi)\cite{le2014tuhoi} dataset is a HOI dataset extended from ILSVRC 2013\cite{ILSVRC15} dataset. Human annotators provide action descriptions with open and word based categories, creating multiple actions with same meaning. The Human Interacting with Common Objects (HICO) \cite{chao2015hico} dataset resolves this issue by incorporating sense-based actions extracted from MS-COCO\cite{coco} image captions. Annotators from AMT label selected images from Flickr for each object category and action keyword. These datasets are not well suited for scene understanding or action recognition within household settings and for human-robot collaboration applications. Furthermore, the above datasets highly depend on the action descriptions from the underlying datasets. In contrast, our dataset has an advantage by procedurally creating unlimited amounts of synthetic data with an option for adding new human poses with novel interactions within indoor scene with minimal efforts.

\subsection{SG Datasets}
The incapacity of current datasets to recognize relations between discrete objects was discovered by Visual Genome \cite{visual_genome} dataset. To tackle this, Visual Genome dataset added a set of descriptions for manually marked multiple image regions. On similar lines, panoptic visual understanding and recognition in open-world is addressed by the AS-1B \cite{as1b} dataset, where a semi-automatic data engine is employed. Large language models (LLMs) and vision-language models (VLMs) are used to generate initial annotations, followed by dataset refinement through human feedback.  Even though it covers variety of relations, it has sparse annotations with less focus on human-object interactions occurring in indoor scenes. 

Earlier scene graph datasets included bounding-box based annotations, 
while leaving out background details. To improve this, scene graph representation in the form of panoptic segmentation is introduced in the PSG \cite{psg} dataset. As opposed to Visual Genome and COCO, predicates from PSG are more concise. The annotation process, however, heavily relies on human annotators and often emphasizes only the most salient relations. Conversely, our dataset offers complete scene graph annotations that are not limited to pre-established image regions. Rather, all relations are derived using a well-defined set of rules, ensuring consistent and reliable annotations.

Haystack \cite{haystack} tackled the problem of incomplete scene graph annotations by including explicit negative ground truth in the dataset, leading to more reliable data. However, the manual annotation process resulted in imprecise predicate definitions. Building upon this idea, our dataset also includes negative ground truth and is densely annotated.

3DSSG\cite{3dssg} and VLA-3D\cite{vla3d} are two scene graph datasets with indoor scenes. 3DSSG is based on 3RScan\cite{3rscan} containing scans of real-world indoor environments, while latter aggregates 3D real-world scans from multiple datasets. 
3DSSG dataset is build by combining semantic annotations and geometric information, by enforcing a broad set of rules for relation extraction. As the underlying scenes are based on real-world scans, they tend to be noisy, hence requiring human verification. Similarly, VLA-3D extracts spatial relations by applying a set of pre-defined rules on object's rotated bounding boxes. 
Our dataset generates much more fine-grained details such as distances, angles in the form of parametric relations as well as human-object interactions, providing reliable ground truth. Additionally, precise annotations are obtained by computing relations based on voxels rather than bounding boxes, giving much more rich annotations as compared to manual annotations.

\begin{figure*}[t!] 
  \centering
  \includegraphics[width=0.7\textwidth]{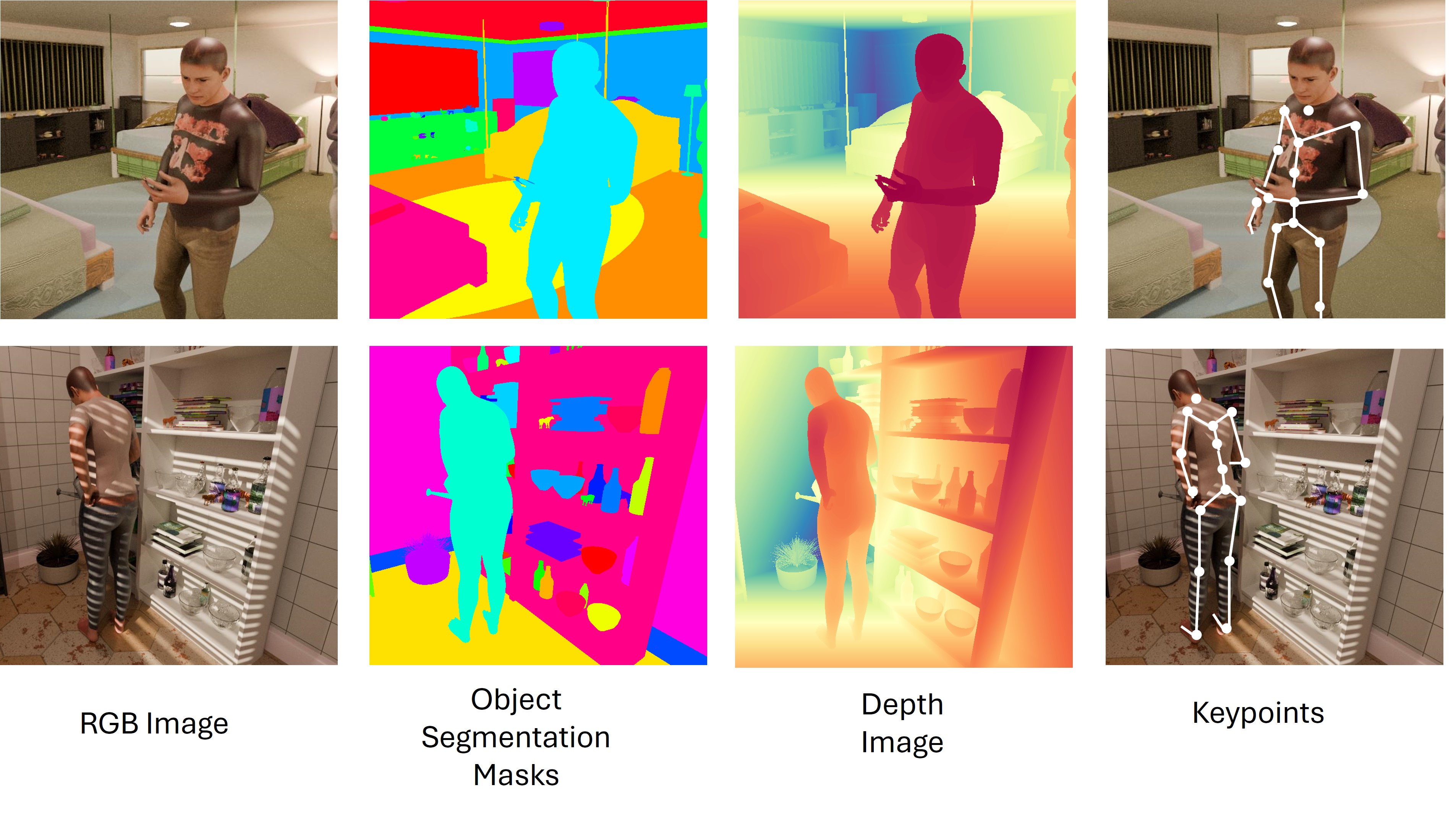}
  \caption{Depicts ground truth data obtained from the scene generation and human insertion pipeline. Ground truth includes realistic RGB images, object segmentation masks, depth images and human keypoints rendered for multiple camera instantiations within the scene.}
  \label{fig:ground_truth}
\end{figure*}





\section{Method}
In this section, we introduce the HOIverse dataset creation pipeline. We explain how the scenes are generated, how we add humans to the scenes, and how we extract ground truth annotations for human-object interactions from the generated scenes.


\subsection{Scene Generation}
We aim to generate a synthetic dataset with human-object interactions, typically observed in household settings for scene graph generation. We include a common set of actions, namely standing, sitting, holding something, touching objects, eating, sleeping, exercising, etc. to make our scene graph dataset usable for human-robot collaboration applications. We follow the approach presented for CoPa-SG \cite{infinisg} and extend their pipeline to include humans that interact with the environment. Similar to \cite{infinisg}, we first employ Infinigen \cite{infinigen2024indoors} to generate high quality procedural indoor scenes.
Infinigen is then also used to render fine grained RGB frames, depth maps, surface normals, and segmentation maps. Other synthetic scene generators\cite{procthor, scenescript, scannet}, generate scenes with predefined elements. On the contrary, Infinigen creates realistic scenes with continuously varying scene objects, adding diversity throughout the dataset.
Fig.\ref{fig:ground_truth} shows two example views with their respective ground truth data. With HOIverse, we additionally provide human keypoints ground truth as compared to CoPa-SG.

For automatic scene graph annotation, we employ the approach described in \cite{infinisg}. Compared to prior scene graph datasets, this allows us to include parametric relations in our dataset. Parametric relations encode an additional parameter for each extracted relation that provide a more fine-grained description.
For example, a \textit{next to} relation can also include a distance parameter that accurately defines the actual distance.
This helps in cases where a binary classification is not sufficient. Human annotators may decide differently if two objects are next to each other or not given their actual distances. Furthermore, applications might have different requirements for various relations. Using parametric relations, an application can directly retrieve the parameter instead of retraining a new network for each new binary definition of a relation. For HOIverse, we include the set of parametric relations from \cite{infinisg} and further extend it by introducing three additional parametric relations including \textit{looking at, pointing at} and \textit{body facing}  while adapting the extraction to all humans in the scene.


\subsection{Human Populator}

This section describes how we insert human meshes with varying textures into existing 3D scenes. The presented approach can be used with any 3D scene, irrespective of the underlying scene generator. 

Even though prior work has made considerable advances in human mesh placement \cite{posa,posegen}, we opt for a more traditional approach when populating the scenes with humans. State-of-the-art neural network-based methods are very flexible, but are still subject to placing errors.
For HOIverse, we strive to generate reliable, correct (i.e. error-free) and precise ground truth data. As such, we prefer a more deterministic approach and rely on a set of predefined poses that are positioned in the scene.

\begin{figure}[h!]
    \centering
    \includegraphics[width=6cm]{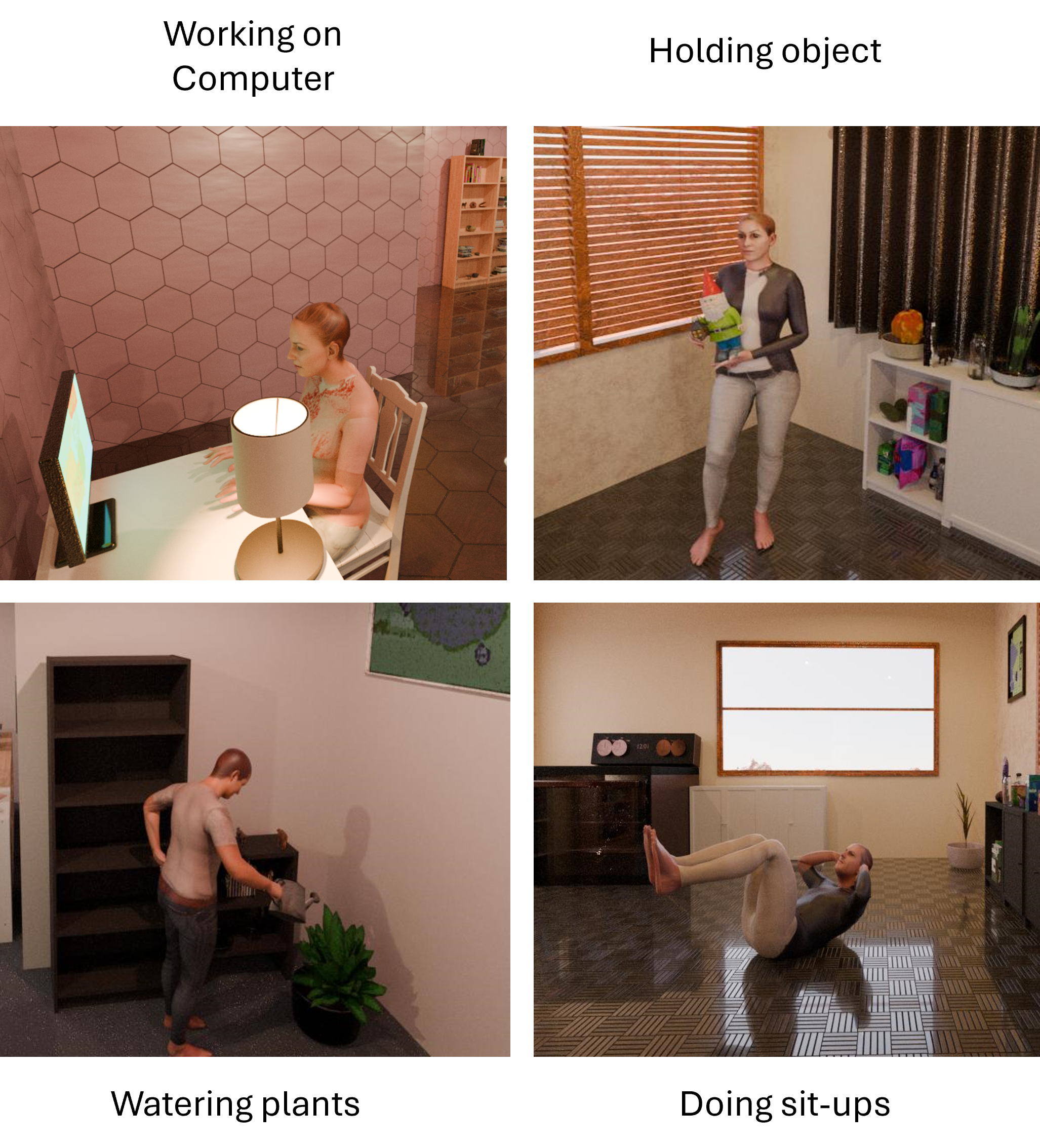}
    \caption{Qualitative examples of supported relation types in HOIverse that involve humans.}
    \label{fig:sample_poses}
\end{figure}

During the human placement step, we sample from a set of custom poses or poses adapted from AGORA\cite{agora} dataset. Since our aim is to insert humans interacting with existing objects within the scene, we generate custom poses such as watering plant or working on computer, which are often difficult to find in any open-source human pose datasets. The required human pose is achieved by adjusting human's joint positions and orientations as per the desired object of interaction.  
The AGORA dataset consist of 4.5k different poses for male, female, and children, obtained by fitting SMPL-X\cite{smplx} model to 3D scans, enabling variation with body shape and gender. Human poses from AGORA include no texture information. Texture information makes rendered RGB images look more realistic as compared to gray or texture-less meshes. Therefore, for imposing variability across our dataset, textures are adapted from SMPLitex \cite{smplitex}. SMPLitex provides 3D human texture dataset created using recent generative models for 2D images. 
Fig.\ref{fig:sample_poses} shows some of the poses included in the dataset.

Infinigen is a procedural scene generation framework providing realistic indoor scenes. We extend it to support human insertion. To do so, we introduce a HumanPopulator pipeline responsible for spawning humans at different locations interacting with scene objects. Currently our dataset supports 14 kinds of human-object interactions, which includes holding
certain object, sitting, watering plants, touching certain objects, looking at, pointing at etc. (see section \ref{sec:dataset_statistics}). We first identify valid unoccupied positions for human insertion by summing over object's bounding box and subtracting it from floor area. Once free space is identified, poses are sampled from set of available poses based on predefined rules. These rules include checking for the presence of desired object of interaction, availability of unoccupied space and type of the room - kitchen or living room. Certain positions like holding glass, holding apple does not require specific positioning within the scene. However, poses like watering plants, sitting on sofa, touching a chair requires pose alignment as per the object orientation. Human poses are further filtered considering object categories within the room, i.e poses like sitting on chair, watering plant are sampled only if chair or plant is present in the scene.  




\subsection{Egocentric Viewpoints}
Egocentric viewpoint definition becomes important to accurately model human-object interactions. For example with such perspective, we can extract a relation describing at which object a human is looking. 

\begin{figure}[h!]
    \centering
    \includegraphics[width=6cm]{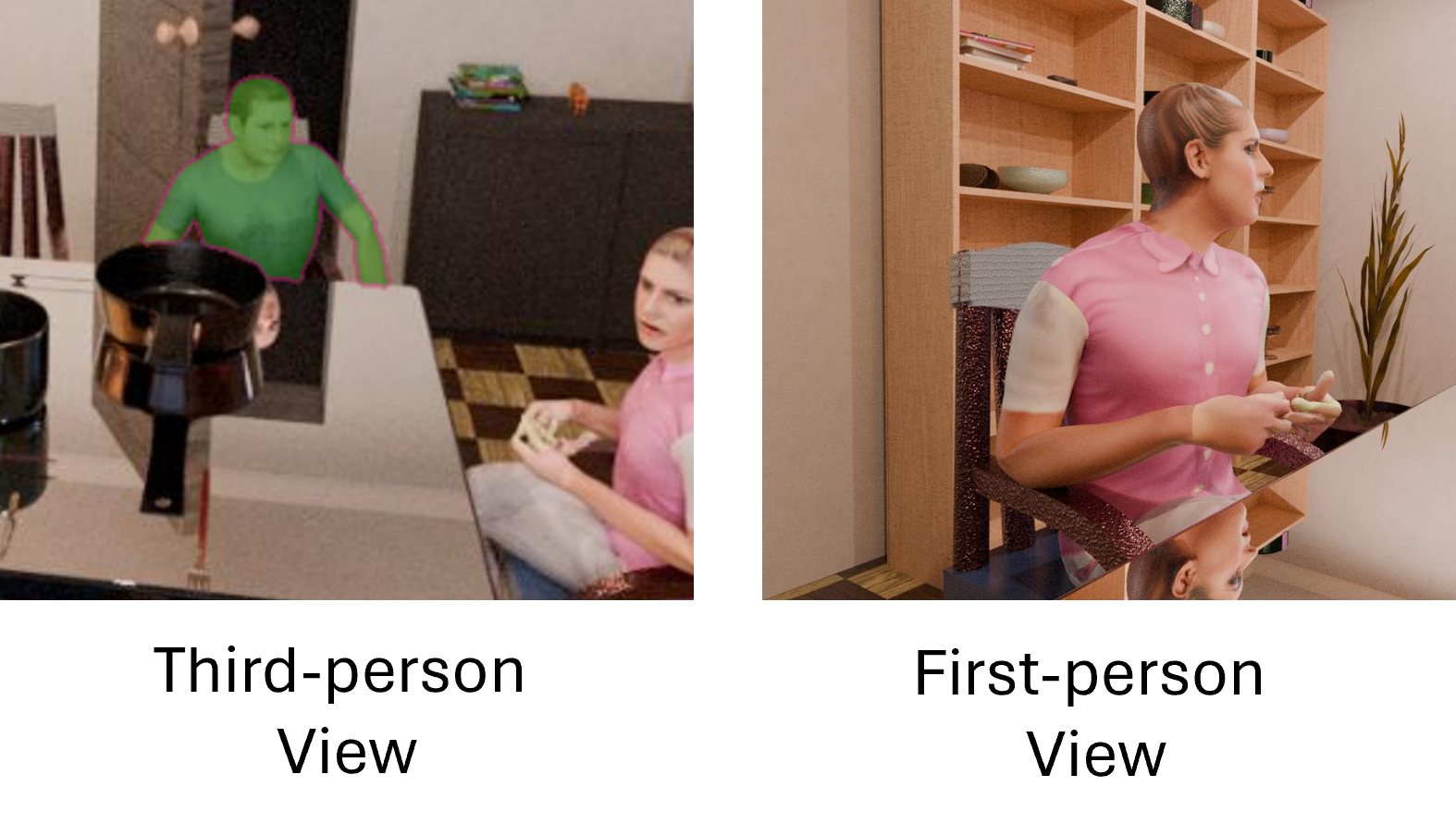}
    \caption{Viewpoint examples: left image shows a third-person view from one of the cameras and the right image shows the first-person view of the highlighted human from the left image.}
    \label{fig:egocentric_view}
\end{figure}

Thus, to enhance the scene graph description and to accurately capture human-object interactions, we assign a dedicated camera for each person in the scene. Its viewpoint is associated with the human's head pose, consequently rendering the ground truth with the "eyes" of the human (first-person view). Positioning of the camera as well as the rendered viewpoint is visualized in Fig.\ref{fig:egocentric_view}.

\subsection{Relation annotation}
\label{sec:relation_annotation}
Existing scene graph datasets lacks the ability of understanding a scene completely by only focusing on salient relations. Similar to CoPa-SG \cite{infinisg}, we compute parametric relations between humans and surrounding objects along with relations describing human-object interactions. This helps to understand the environment and the reason for an action.
The annotation file describing human-object interactions is generated procedurally at the instance of generation of every human-populated 3D scene. In certain object interactions, a human may perform multiple actions at the same time. For instance, \textit{human is eating food while sitting on the chair}, such interactions are divided into two relations, (\textit{human, sitting on, chair}) and (\textit{human, eating, food}). Additionally, spatial relations, like \textit{left, right, front, behind, next to} are computed for both camera-independent as well as camera-dependent views by incorporating parametric relations from \cite{infinisg}. Camera-independent relations are computed for all object instances in the scene, where spatial relations are defined according to the object orientation with respect to the world coordinate system. On the other hand, camera-dependent relations are defined relative to the camera's orientation within the world. Human-object interactions are described for the entire scene and later filtered considering the visibility of both human and object within the camera frame.


\begin{figure}
    \centering
    \includegraphics[width=0.5\linewidth]{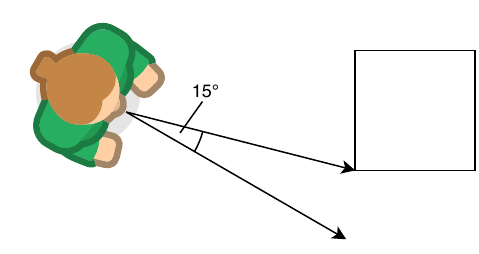}
    \caption{Schematic of the \textit{looking at} relation. Using a ray cast approach, a vector is extracted that connects a human and another object. The vector with the smallest possible deviation from the viewing direction is chosen for the final relation.}
    \label{fig:lookingat}
\end{figure}

\begin{figure}[h!]
    \centering
    \includegraphics[width=8cm]{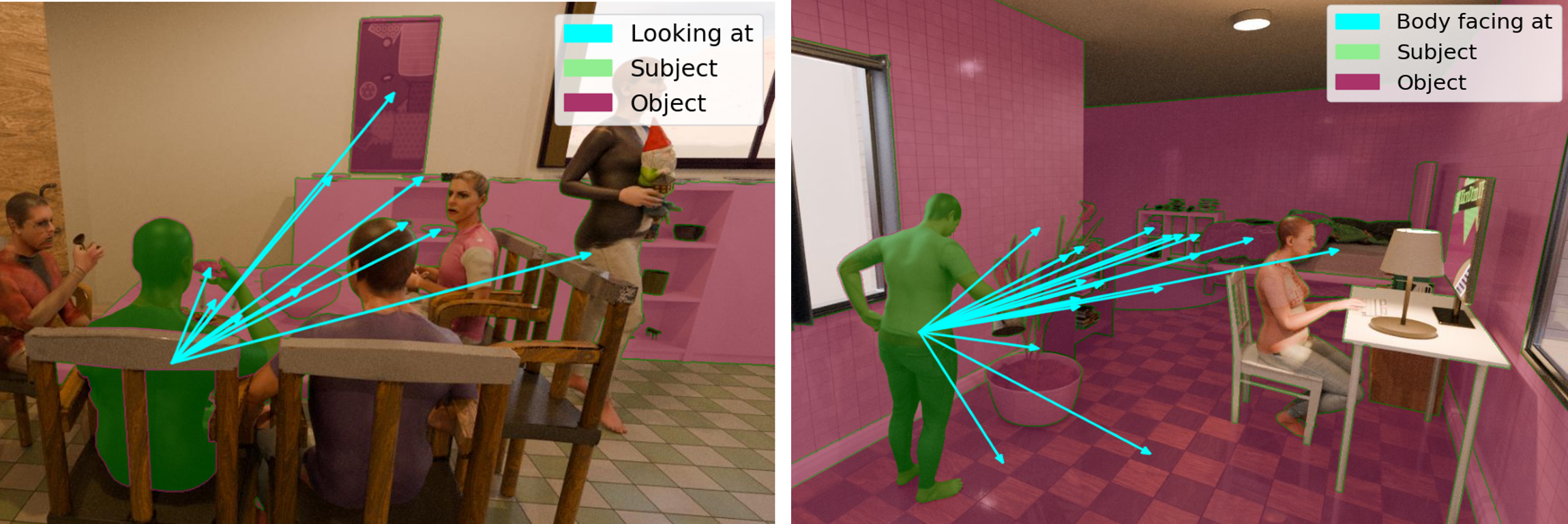}
    \caption{Example annotations for \textit{looking at} and \textit{body facing} relations provided in HOIverse dataset.}
    \label{fig:looking_at}
\end{figure}

In addition, we include \textit{looking at}, \textit{pointing at}, and \textit{body facing} (see Fig.\ref{fig:looking_at}) relations that depend on the pose of the participating human.
The \textit{looking at} relation determines whether a person looks at an object or not. To extract the relation from the underlying scene, we locate the head joint position for each human in the scene and follow the view direction until it hits an object. This is implemented using ray casts in Open3D \cite{open3d}.
Following \cite{infinisg}, we extract a parametric variant of the \textit{looking at} relation. We sample additional rays that deviate from the main viewing direction and keep track which objects are being hit.
If an object is hit by multiple rays, we choose the ray with the smallest deviation from the main viewing direction.
This deviation is stored in the ground truth as a parametric relation. The maximum allowed deviation is 90 degrees; everything above is not regarded as \textit{looking at}.
For the \textit{pointing at} relation, we use a very similar approach. Instead of the head joint, we use the left and right index finger joints to get a pointing direction.
Again, we sample additional rays to retrieve parametric relations.
Finally, we extract a \textit{body faces} relation which determines the direction in which the torso of a person faces. We use the \textit{spine3} joint from SMPL-X to determine the orientation of the torso and apply the same procedure as with the \textit{looking at} and \textit{pointing at} relations.

\subsection{Keypoints}

In addition to scene graph annotations, we also include 2D keypoint annotations of human body joints for each rendered image.
Since we are using SMPL-X \cite{smplx} meshes to represent humans in the scenes, we provide body shape and body pose parameters for each rendered scene. Additionally, we project each joint position to camera coordinates for 2D keypoint annotations. Since we are using SMPL-X, the keypoints also contain finger positions.

\section{Results}
In this section, we provide statistics of the extracted HOIverse dataset and use it to train several scene graph generation models to analyze their performance on HOIverse.

\subsection{Dataset Statistics}
\label{sec:dataset_statistics}

Our HOIverse dataset consists of ground truth RGB images, segmentation masks, depth images, human keypoints along with scene graph annotations per rendered image. 
We have generated over 525 indoor scenes procedurally using Infinigen, where each scene has on average 31 camera views,
accumulating around 16.2k rendered images in our HOIverse dataset. The rendered ground truth data is acquired from various view points within the scene, including human's egocentric viewpoint as well. Currently, the HOIverse dataset provide 14 types of object interactions in total as mentioned in Table \ref{tab:hoi}. We are continuously enhancing our HOIverse dataset to include more number of interactions. 

Each scene from the HOIverse dataset is populated with 2-6 humans interacting with the scene, resulting into total 40k interaction annotations for all human-object interactions listed in Table \ref{tab:hoi}.
For each of the RGB image, we accumulate spatial as well as human object relations, resulting on an average in 1.35k relations per viewpoint and 22M scene graph annotations for the entire HOIverse dataset.  The entire dataset is divided into train, validation and test dataset, where validation set and test set constitutes 20\% and 10\% of the whole dataset respectively. While creating the data split, all rendered images per scene are aggregated in the same data split.


\begin{table}[h!]
\begin{tabular}{|
>{\columncolor[HTML]{C0C0C0}}l |l|}
\hline
\begin{tabular}[c]{@{}l@{}}Human \\ Object\\ Interactions\end{tabular} & \begin{tabular}[c]{@{}l@{}}holding object, sitting, standing on floor, waving, \\watering plant, touching object, doing sit-ups on floor, 
\\ working on computer, sleeping, checking time on watch, \\ eating, looking at, pointing at, body facing \end{tabular} \\ \hline
\end{tabular}
\caption{List of human object interactions included in HOIverse dataset.}
\label{tab:hoi}
\end{table}

\subsection{Scene Graph model evaluation}

We perform predicate classification (\textit{PredCls}) with existing state-of-the-art scene graph generation models including  MotifNet\cite{motifs}, VCTree\cite{VCTree} and DSFormer\cite{dsformer} on our HOIverse dataset. Out of these, DSFormer is a two-stage method, where object detection and predicate classification is carried out separately. This eliminates the effect of object detection performance on relation prediction. 
We follow the approach employed in CoPa-SG dataset\cite{infinisg} to train on parametric relations as well as human-object interactions. 

Traditional scene graph generation models like MotifNet and VCTree use binary labels for predicate classes and thus, are not suitable in predicting parameters along with predicate classes. Therefore, parameter thresholding is performed to categorize parametric relations into distinct spatial relations. We do it in similar fashion as CoPa-SG, where relations are split into positive and negative classes based on threshold coefficient. Relations with an angle parameter $\leq 10^\circ$ are considered as positive annotations, whereas angle parameter $\geq 20^\circ$ are denoted as negative annotations. For distance parameter, values $\leq$ 1\textit{m} and $\geq$ 1.2\textit{m} are marked as positive and negative respectively.

As the HOIverse dataset is exhaustively annotated, evaluation metrics used in traditional scene graph generation models are not useful.
As traditional scene graph datasets are sparsely annotated, Recall@k and Mean Recall@k (\textit{mR@k}), where $k \in \{20, 50, 100\}$ metrics are primarily used. Because our dataset is densely annotated, the models could only achieve 0.020 and 0.032 \textit{mR@50} and \textit{mR@100} respectively with such low \textit{k} values. Also, \textit{mR@k} is computed when subject-object pair has a single predicate, therefore, it is not optimal for our dataset due to multi-label subject-object instances. Therefore, we use No Graph Constraint Mean Recall@k (\textit{ng-mR@k}), and we compute it over 1000 predicate relations. Furthermore, we employ the Mean Average Precision (\textit{mAP}) metric computed on all predicates, as our dataset supports positive as well as negative relation classification. Table \ref{tbl:model_performance} summarizes the model performances on our HOIverse dataset. DSFormer has better performance as compared to MotifNet and VCTree, demonstrating the importance of \textit{mAP} metric for model evaluation.
\begin{table}[h!]
    \begin{center}
        \begin{tabular}{lrrrr}
            \toprule
            & \multicolumn{2}{c}{PredCls $\uparrow$} \\
            \cmidrule(lr){2-3}\cmidrule(lr){4-5}
            Method & mAP & ng-mR\\
            \midrule
            MotifNet \cite{motifs} & 0.504 & 0.312 \\
            VCTree \cite{VCTree} & 0.511 & 0.315\\
            DSFormer \cite{dsformer} & 0.667 & 0.218\\
            \bottomrule
        \end{tabular}
    \end{center}
    \caption{Model Performance on HOIverse dataset evaluated on predicate classification (\textit{PredCls}). These models are trained on our dataset including \textit{front, behind, above, below, left, right, on, next-to, touching, looking at, pointing at, body facing, sitting} and \textit{holding} relations.}
    \label{tbl:model_performance}
\end{table}


\begin{figure}[h!]
    \centering
    \includegraphics[width=6cm]{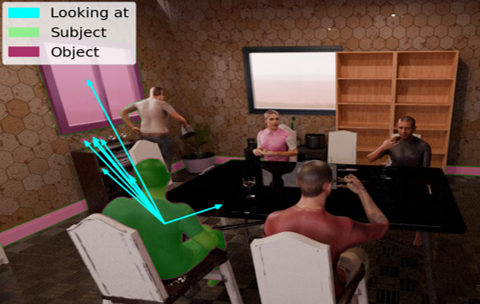}
    \caption{Visualization for model prediction, where \textit{looking at} relation is depicted here from all predicted relations.}
    \label{fig:scene_image}
\end{figure}

\section{Conclusion}

We presented the novel HOIverse dataset, a dataset at the intersection of scene graph and human-object interactions. This dataset is created to focus on all aspects of the scene rather than just relying on salient relations for better visual scene understanding. To eliminate the effects of sparse and inconsistent annotations due to manual labeling, we incorporate the approach employed in CoPa-SG to generate complete and consistent annotations. We extended their work by including humans in the scene to provide a reliable dataset for scene understanding in human-robot collaboration. We designed our own pipeline for integrating humans in different poses interacting with the scene objects and computing novel parametric relations including \textit{looking at, pointing at} and \textit{body facing} for each human instance. This helps to generate unlimited synthetic data to fuel the research in the field of scene understanding.
With the help of our dataset, we aim to make the task of complete scene understanding simpler and to accelerate further research in the field of human-robot collaboration.






\bibliographystyle{IEEEtran}
\bibliography{IEEEabrv,references}

\end{document}